\documentclass{article}
\usepackage[utf8]{inputenc}
\usepackage[T1]{fontenc}
\usepackage{arxiv}
\usepackage{graphicx}
\usepackage{microtype}

\usepackage{multirow}
\usepackage{mathtools}
\usepackage{amssymb}
\usepackage{amsthm}
\usepackage{mathrsfs}
\usepackage[title]{appendix}
\usepackage{xcolor}
\usepackage{textcomp}
\usepackage{manyfoot}
\usepackage{booktabs}
\usepackage{algorithm}

\usepackage{algpseudocode}
\usepackage{listings}

\usepackage{natbib}
\usepackage{url}
\usepackage[space]{grffile}
\usepackage{bm}
\usepackage{nicefrac}
\usepackage{array}
\usepackage{varwidth}
\usepackage{caption}
\usepackage{subcaption}
\usepackage{tikz}
\usetikzlibrary{positioning} 
\usetikzlibrary{arrows}
\usetikzlibrary{calc}
\usepackage[colorlinks=true,allcolors=blue,hypertexnames=false,pdftitle={Bayesian Flow Networks for Offline Trajectory Planning},pdfauthor={Ludvig Killingberg and Helge Langseth}]{hyperref}

\newcommand{\lfm}[0]{BFN-RL}

\newcommand{\vx}{\bm{x}}
\newcommand{\Vx}{\mathbf{x}}
\newcommand{\vy}{\bm{y}}
\newcommand{\Vy}{\mathbf{y}}

\newcommand{\Vex}{\mathbf{e_x}}
\newcommand{\Vea}{\mathbf{e}_a}
\newcommand{\vtheta}{\bm{\theta}}
\newcommand{\vomega}{\bm{\omega}}
\newcommand{\vmu}{\bm{\mu}}
\newcommand{\N}{\mathcal{N}}
\newcommand{\R}{\mathbb{R}}
\newcommand{\E}{\mathbb{E}}
\newcommand{\I}{\bm{I}}
\newcommand{\known}{\text{known}}
\newcommand{\unknown}{\text{unknown}}

\newcommand{\defeq}{\overset{\text{\tiny def}}{=}}

\title{Bayesian Flow Networks for Offline Trajectory Planning}
\author{Ludvig Killingberg \\
Norwegian University of Science and Technology \\
\texttt{ludvig.killingberg@ntnu.no}
\And
Helge Langseth \\
Norwegian University of Science and Technology}
\date{}
\renewcommand{\shorttitle}{BFN-RL}
\renewcommand{\headeright}{Preprint}
\renewcommand{\undertitle}{Preprint}

\begin{document}
\maketitle

\begin{abstract}

Offline reinforcement learning (RL) leverages static datasets to learn decision policies without real-time environment interaction. While recent sequence-modeling approaches rely on continuous diffusion models for trajectory synthesis, applying these methods to discrete planning tasks requires a categorical formulation rather than the standard Gaussian construction. We present BFN-RL, a unified generative modeling framework for offline RL based on Bayesian Flow Networks (BFNs). By iteratively evolving distribution parameters rather than noisy data instances, BFN-RL natively models both discrete and continuous trajectory spaces within a single probabilistic formulation. The categorical planner generates future state sequences, and a learned inverse-dynamics model converts consecutive generated states into actions. Evaluations in discrete planning and continuous control show that BFN-RL can generate effective trajectories across both categorical and continuous state spaces. Our results establish BFNs as a versatile generative foundation for offline trajectory planning across data modalities.
\end{abstract}

\section{Introduction}\label{introduction}
Offline reinforcement learning (RL) \citep{sutton2018reinforcement, levine2020offline} is a powerful paradigm that leverages static, previously collected datasets to learn effective decision policies without requiring real-time environment interaction. By eliminating the safety hazards associated with online exploration, offline RL is particularly suited for high-risk domains such as autonomous driving and medical decision-making. In recent years, framing offline RL as a conditional sequence-modeling task \citep{janner2021sequence, chen2021} has emerged as a compelling alternative to traditional value-based methods, which frequently suffer from value overestimation on out-of-distribution state-action pairs \citep{agarwal2020, levine2020offline}. By viewing trajectory generation through the lens of conditional generative modeling, sequence-based agents synthesize high-return trajectories by capturing complex temporal dependencies across long horizons \citep{janner2022diffuser, ajay2022conditional}.

Despite their empirical success, contemporary sequence-modeling approaches rely almost exclusively on Gaussian Denoising Diffusion Probabilistic Models (DDPMs) \citep{ho2020denoising, janner2022diffuser, ajay2022conditional}. While diffusion models excel in continuous control domains, discrete states and actions require a categorical formulation rather than the standard Gaussian construction~\citep{austin2023structureddenoisingdiffusionmodels, lou2024discretediffusionmodelingestimating}. This motivates studying a generative framework that handles both data types natively.

To address this limitation, we present \emph{BFN-RL}, an offline reinforcement learning framework grounded in Bayesian Flow Networks (BFNs) \citep{graves2024bayesian}. Unlike diffusion models that iteratively denoise corrupted data instances, BFNs operate by updating the parameters of an input distribution via Bayesian inference driven by continuous-time parameter flows \citep{graves2024bayesian}. This formulation yields a unified generative paradigm that natively handles categorical, continuous, and discretized variables within the same probabilistic framework.

Our main contributions are summarized as follows:
\begin{itemize}
\item \textbf{Unified Generative Planning Framework:} We introduce BFN-RL, establishing a parameter-flow paradigm for offline sequence modeling that operates seamlessly across both discrete and continuous state-action spaces \citep{graves2024bayesian}.

\item \textbf{Categorical Trajectory Planning:} We combine a categorical BFN state-sequence model with learned inverse dynamics to obtain a discrete planner.
\item \textbf{Cross-Domain Evaluation:} We demonstrate that the same BFN planning formulation is viable in both discrete and continuous control problems.
\end{itemize}

\section{Preliminaries}\label{preliminaries}

\subsection{Reinforcement Learning}\label{reinforcement-learning}
Reinforcement learning (RL) is a framework for learning to make
decisions in an environment~\citep{sutton2018reinforcement}. 
The interactions with the environment are modeled as a Markov decision process (MDP), which is a tuple
$(\mathcal{S}, \mathcal{A}, \mathcal{P}, \mathcal{R}, \gamma)$, where
\(\mathcal{S}\) is the state space, \(\mathcal{A}\) is the action space,
\(\mathcal{P}\) is the transition function, \(\mathcal{R}\) is the
reward function, and \(\gamma\) is the discount factor. 
In an environment where the agent performs the action $a\in\mathcal{A}$ in state \(s\in\mathcal{S}\), the next state, \(s^\prime\in\mathcal{S}\) is sampled from $\mathcal{P}(s, a)$, i.e., is only dependent on the current state and action, not the history of previous states and actions. 
In other words, the domain adheres to the Markov property.
Let \(r_t\) denote the reward received at time \(t\), and let 
$R_t = \sum_{i=0}^{\infty} \gamma^i r_{t+i}$ be the discounted cumulative reward obtained starting from time $t$. 
Now, one goal of RL is to learn a policy \(\pi: \mathcal{S} \rightarrow \mathcal{A}\) that maximizes the expected return \(\E[R_t]\), where the expectation is taken over the uncertainty defined by both the transition function and the stochastic strategy $\pi$.

The exploration-exploitation trade-off is a fundamental challenge in reinforcement learning, typically associated with online learning scenarios where agents iteratively interact with an environment to learn optimal policies. 
Exploration involves sampling actions to gather information about the environment, potentially leading to the discovery of better strategies, while exploitation entails leveraging known information in an attempt to maximize the expected returns. 
Much of the research in online RL is dedicated to striking a balance between exploration and exploitation, devising algorithms that effectively navigate this trade-off to converge to optimal or near-optimal policies.

\subsection{Offline RL}\label{offline-rl}
In the realm of offline reinforcement learning, the primary objective is to learn effective policies from a fixed dataset, without
the need for online interactions~\citep{levine2020offline}. 
In this context, where 
the exploration aspect is inherently absent, the focus shifts towards effectively utilizing the available dataset to optimize policies. 
Traditionally, RL has been concerned
with estimating stationary policies or single-step models, leveraging
the Markov property to factorize problems over time. 
However, applying standard RL methods to offline settings is challenging because methods relying on an estimated value function often suffer from overestimating the value of out-of-distribution state-action pairs.
Various methods have been proposed to address this issue, including constraining the policy to be close to the data distribution~\citep{peters2010relative} or
using a conservative value function~\citep{kumar2020conservative}.
Our solution, on the other hand, is to produce a sequence of steps that will be generated conditionally on the objective of the RL agent.   
An intriguing perspective emerges when we view RL through the lens of
sequence modeling. 
Instead of treating it as a specialized domain, we can consider RL as a generic sequence modeling problem. 
The crux of this viewpoint lies in producing a sequence of actions that leads to a sequence of high rewards. 
Earlier work has solved this by conditioning the model on returns such that trajectories with high returns can be generated in online settings~\citep{ajay2022conditional,janner2021sequence}. 
By adopting this perspective, we can simplify design decisions and dispense with many components commonly found in offline RL algorithms. 
This approach not only demonstrates flexibility across various tasks such as long-horizon dynamics prediction, imitation learning, goal-conditioned RL, and offline RL but also yields state-of-the-art planners in sparse-reward, long-horizon
scenarios~\citep{janner2022diffuser,ajay2022conditional}.

\subsection{Denoising Diffusion Probabilistic Models}\label{diffusion-probabilistic-models}
Since the current state of the art in this domain~\citep{janner2022diffuser,ajay2022conditional}
rely on diffusion models as the generative model, we will give a brief introduction to denoising diffusion probabilistic models (DDPMs)~\citep{ho2020denoising}.
This will also serve as a backdrop for our discussion of Bayesian flow networks, which follows in Section \ref{bayesian-flow-networks}. 
DDPMs are a type of generative model inspired by non-equilibrium thermodynamics. 
The model is defined by a \textit{forward process} that slowly adds Gaussian
noise to data, and its \textit{reverse}, that amounts to learning to iteratively denoise the noisy data. 
Diffusion models have primarily been used for image generation, but have
also shown state-of-the-art performance in other domains, like video generation and 3D model~\citep{ho2022video,luo2021diffusion}.

Given data \(\Vx_0 \sim q(\Vx)\), we define the \emph{forward process} to produce a sequence of noisy samples \(\Vx_1, \dots, \Vx_K\),
\begin{equation*}
	q(\Vx_k \vert \Vx_{k-1}) = \N (\Vx_k; \sqrt{1 - \beta_k} \Vx_{k-1}, \beta_{k} \I).
\end{equation*}
where \(\{\beta_i \in (0,1)\}_{i=1}^K\) is a carefully chosen variance schedule. 
A nice property of the forward process is that we can directly sample \(\Vx_k\) at any step \(i\), because the distribution \(q(\Vx_k \vert \Vx_0)\) can be derived using the property that a sum of uncorrelated normally distributed random variables is normally distributed. 
Let \(a_k = 1 - \beta_k\) and \(\overline{a}_k = \prod_{i=1}^k a_k\), then
\begin{equation*}
	q(\Vx_k \vert \Vx_0) = \N \left(\Vx_k; \sqrt{\overline{a}_k} \, \Vx_0, \left(1 - \overline{a}_k \right) \I \right).
\end{equation*}
Note also that
\begin{equation}
\label{eq:reverse_conditioned}
q(\Vx_{k-1} \vert \Vx_k, \Vx_0) = \N \left( \Vx_{k-1} ; \tilde{\bm{\mu}}(\Vx_k, \Vx_0), \tilde{\beta}_k \I \right),
\end{equation}
where 
\begin{equation}\label{eq:bfn_mu}
\tilde{\bm{\mu}}(\Vx_k, \Vx_0) = \frac{\sqrt{\overline{a}_{k-1}}\beta_k}{1 - \overline{a}_k} \Vx_0 + \frac{\sqrt{a_k}(1 - \overline{a}_{k-1})}{1 - \overline{a}_{k}}\Vx_k, 
    \qquad \tilde{\beta}_k = \beta_k \frac{1-\overline{a}_{k-1}}{1 - \overline{a}_k}.
\end{equation}

While the forward process creates a noisy representation of data, the \emph{reverse process} aims to iteratively recreate samples from noise by modeling and then sampling from \(q(\Vx_{k-1} \vert \Vx_k)\). Let \(p_\theta(\Vx_{k-1} \vert \Vx_k)\) be a parameterized approximation of \(q(\Vx_{k-1} \vert \Vx_k)\). This means that we define a neural network model with trainable parameters $\boldsymbol{\theta}$ that  outputs $\bm{\mu}_\theta(\Vx_k, k)$ and $\Sigma_\theta(\Vx_k, k)$ so that
\begin{equation*}
	p_\theta(\Vx_{k-1} \vert \Vx_{k}) = \N\left( \Vx_{k-1} ; \bm{\mu}_\theta(\Vx_k, k), \Sigma_\theta(\Vx_k, k) \right).
\end{equation*}
\citet{ho2020denoising} chose to fix the variance term \(\bm{\Sigma}_\theta(\Vx_k, k)\) as a constant \(\sigma_k^2 = \tilde{\beta}_k\), see Eq.~\eqref{eq:bfn_mu}.
We therefore only look at how \(\bm{\mu}_\theta(\Vx_k, k) \) is estimated. 
First, we consider the identity 
\begin{equation*}
\bm{\tilde{\mu}}_k(\Vx_k,\Vx_0) = \frac{1}{\sqrt{a_k}} \left( \Vx_k - \frac{1-a_k}{\sqrt{1 - \overline{a}_k}} \bm{\epsilon}_k \right), 
\end{equation*}
where \(\bm{\epsilon}_k \sim \N(0,\I)\); 
cf. Eqs.~\eqref{eq:reverse_conditioned} and \eqref{eq:bfn_mu}. 
Since \(\Vx_k\) is known during training, we can choose to predict \(\bm{\epsilon}_k\), rather than \(\bm{\tilde{\mu}}_k\) directly. 
Empirically, this has shown better results. 
Let us define \(\bm{\epsilon}_\theta(\Vx, k)\) as a model that predicts the noise, \(\bm{\epsilon}_k\). 
This means that we can define \(\bm{\mu}_\theta(\Vx_k, k) = \frac{1}{\sqrt{a_k}} \left( \Vx_k - \frac{1-a_k}{\sqrt{1 - \overline{a}_k}} \bm{\epsilon}_\theta(\Vx_k, k) \right)\). 
\citet{ho2020denoising} derive the following loss function to minimize the difference between \(\bm{\mu}_\theta\) and \(\bm{\tilde{\mu}}\):
\begin{equation*}
	L(\theta) = \mathbb{E}_{k \sim [1,K], \Vx_0, \bm{\epsilon}_k} \left[ \frac{\beta_k^2}{2\sigma^2_k a_k(1 - \overline{a}_k)} \left\lVert \bm{\epsilon}_k - \bm{\epsilon}_\theta(\Vx_k, k) \right\rVert^2  \right].
\end{equation*}

They also present the following simplified loss function that turns out to give better empirical results:
\begin{equation*}
	L(\theta) = \mathbb{E}_{k \sim [1,K], \Vx_0, \epsilon_k} \left\lVert \epsilon_k - \epsilon_\theta(\Vx_k, k)  \right\rVert^2.
\end{equation*}

\subsection{Guided Diffusion}\label{guided-diffusion}
We will discuss three ways diffusion models can condition on variables.
The first, \textit{classifier-guided} diffusion~\citep{dhariwal2021diffusion}, takes as its starting point that we have a trained probabilistic classifier 
that classifies objects $\Vx_k$ during the reverse process.
We want to use this to produce an object $\Vx_0$ that is classified as belonging to a predefined class $y$.
The approach uses the gradients of this classifier's allocated log-likelihood to the class $y$ wrt.\ $\Vx_k$ to ``push'' the reverse process towards objects that are aligned with the conditioning information. 
This method has the advantage that the diffusion model does
not have to be trained with conditioning variables, the guidance is only related to the reverse and only needs the classifier to be trained on conditioning information. 
A model predictor \(\overline{\bm{\epsilon}}_\theta\), guided by a classifier \(h(y \vert \Vx_k, k)\) meant to estimate the probability that the noisy datapoint \(\Vx_k\) belongs to class \(y\), would assume the following form:
\begin{equation*}
	\overline{\bm{\epsilon}}_\theta(\Vx_k, k, y) = \bm{\epsilon}_\theta(\Vx_k, k) - w\sigma_k \nabla_{\Vx_k} \log{h(y \vert \Vx_k, k)},
\end{equation*}
where \(w\) is a hyper-parameter controlling the strength of the guidance.

Secondly,
\textit{classifier-free guidance}~\citep{Ho2022ClassifierFreeDG}, plugs
the conditioning variable
directly into the denoising network as an auxiliary input variable
during training. 
At test time, the auxiliary variable can be set to the conditioning value. 
In this setting, the model predictor takes the following form:
\begin{equation*}
	\bm{\tilde{\epsilon}}(\Vx_k, k, y) = (w+1)\bm{\epsilon}_\theta(\Vx_k, k, y) - w\bm{\epsilon}_\theta(\Vx_k, k),
\end{equation*}
where we again use  \(w\) to denote the hyper-parameter that controls the strength of the guidance.
Classifier-free guidance has shown better practical
performance than classifier-guided diffusion~\citep{Ho2022ClassifierFreeDG}.

Finally, we can also employ inpainting~\citep{lugmayr2022repaint} to
condition on partial observations. 
In the context of image generation, this implies conditioning on some
pixels within the image. 
Consider an image \(\Vx_0\) divided into known pixels \(\Vx_0^{\known}\) and unknown pixels \(\Vx_0^{\unknown}\), and a mask \(\bm{m}\) defining which pixels are known. 
During the reverse process, we define
\begin{equation*}
\Vx_{k-1}^{\known} = \sqrt{\overline{a}_k}\Vx_0 + \sqrt{1 - \overline{a}_k} \bm{z}, \qquad \bm{z} \sim \N (\bm{0}, \mathbf{I}),
\end{equation*}
i.e., $\Vx_{k-1}^{\known}$ is chosen equal to what the forward process would have produced had it started from the known parts of the image.
The unknown pixels at step \(k\), \(\Vx_{k-1}^{\unknown}\), are computed in standard fashion:
\begin{equation*}
	\Vx_{k-1}^{\unknown} = \frac{1}{\sqrt{a_k}} \left( \Vx_k - \frac{\beta_k}{\sqrt{1-\overline{a}_k}} \bm{\epsilon}_\theta (\Vx_k, k) \right) + \sigma_k \bm{z}, \qquad
	\bm{z} \sim \N (\bm{0}, \mathbf{I}).
\end{equation*}
Finally, we have:
\begin{equation}
\label{equ:inpainting_total}
	\Vx_{k-1} = \bm{m} \odot \Vx_{k-1}^{\known} + (1 - \bm{m}) \odot \Vx_{k-1}^{\unknown},
\end{equation}
where \(\odot\) is elementwise multiplication.

\subsection{Bayesian Flow Networks}\label{bayesian-flow-networks}
While there are variations of diffusion models that model discrete data~\citep{lou2024discretediffusionmodelingestimating,austin2023structureddenoisingdiffusionmodels}, these models are not considered state of the art when it comes to generating high-quality discrete data \citep{graves2024bayesian}. 
In an attempt to remedy this shortcoming, \citet{graves2024bayesian} introduced
\textit{Bayesian flow networks} (BFNs), a novel generative model capable of generating
continuous, discrete, and discretized data. 
BFNs resemble diffusion models in that they generate data in an iterative process. Unlike
diffusion models, however, the BFN analogy to the diffusion models' reverse process iteratively evolves \textit{distribution parameters}, not noised versions of data. 
The high-level idea is to start from a prior distribution and iteratively
update the distribution conditioned on a data point sampled from a noisy version of the previous distribution. 
BFNs have been shown to perform well on discrete data \citep{graves2024bayesian}, and are therefore a more natural choice than diffusion models for planning in discrete state spaces. 

Figure~\ref{fig:bfn} illustrates the idea of Bayesian flow networks. 
At each step \(i\), the parameters of a distribution (\(\vtheta_i\)) are updated with noisy samples from the data, \(\vy_i\). 
The level of added noise decreases for each step and is at step \(i\) dictated by the accuracy \(\alpha_i\). The parameters \(\vtheta_i\) are defined as the Bayesian update of the parameters at the previous step, \(\vtheta_{i-1}\), with observation noise parameterized by \(\alpha_i\) using a predetermined update rule \(h(\cdot)\).
While each evolved distribution is distinct from the data distribution, the idea is that the compound of all generated distributions should approximate the data distribution. To sample from the data distribution, an initial uninformative distribution is iteratively updated, gradually concentrating around a single data point. Once the distribution has evolved sufficiently, a single sample from this nearly degenerate distribution will closely approximate a sample from the data distribution.

\begin{figure}[htbp]
	\center
    \scalebox{.75}{

	\begin{tikzpicture}[>=stealth, node distance=2cm, auto]

    \node (theta1) at (0,0) {$\vtheta_{i} \defeq h(\vtheta_{i-1}, \vy_{i-1}, \alpha_{i-1})$};
    
    \node (pF) [left=1cm of theta1] {$\dots$};
    \node (theta0) [left=1cm of pF] {$\vtheta_{0}$};
    \node (pF2) [right=2.5cm of theta1] {$\vtheta_{i+1} \defeq h(\vtheta_i, \vy_i, \alpha_i)$};
    \node (theta2) [right=1cm of pF2] {$\dots$};
    \node (theta3) [right=1cm of theta2] {$\vtheta_{N}$};
    
    \draw[->] (theta0) -- (pF);
    \draw[->] (pF) -- (theta1);
    \draw[->] (theta1) -- (pF2);
    \draw[->] (pF2) -- (theta2);
    \draw[->] (theta2) -- (theta3);
    \end{tikzpicture}
    }
\caption{Generative process for Bayesian flow networks.}\label{fig:bfn}
\end{figure}

\begin{figure}[htbp]
\center
\scalebox{.75}{
\begin{tikzpicture}[>=stealth, node distance=2cm, auto]

    \node (theta1) at (0,0) {$\vtheta_{i} \defeq h(\vtheta_{i-1}, \vy_{i-1}, \alpha_{i-1})$};
    
    \node (pF) [left=1cm of theta1] {$\dots$};
    \node (theta0) [left=1cm of pF] {$\vtheta_{0}$};
    \node (pO) [below=2cm of theta1] {$p_O$};
    \node (pR) [below=2cm of pO] {$p_R$};
    \node (pF2) [right=2.5cm of theta1] {$\vtheta_{i+1} \defeq h(\vtheta_i, \vy_i, \alpha_i)$};
    \node (theta2) [right=1cm of pF2] {$\dots$};
    \node (theta3) [right=1cm of theta2] {$\vtheta_{N}$};

    \draw[->] (theta1) to (pF2);

	\draw[->] (theta0) -- (pF);
    
   	\node (pS) [right=2cm of pR] {$p_S$};
   	\node (x) at (pS|-pO) {$\vx$};

    \draw[->] (theta1) -- node[midway, name=nn, anchor=center, fill=white] {\footnotesize Neural Net$(\vomega)$} (pO);
    \draw[->] (pO) -- node[midway, anchor=center, fill=white, sloped, circle, draw, inner sep=0pt, name=p1] {\small$+$} (pR);
    \draw[->] (x) -- node[midway, anchor=center, fill=white, sloped, circle, draw, inner sep=0pt, name=p2] {\small$+$} (pS);
    \draw[->] (pS) -| (pF2);
    \node (sample)[fill=white] at (x-|pF2) {{\footnotesize Sample} $\vy_i$};
    
    \node (noise)[inner sep=1] at ($(p1)!0.5!(p2)$) {\footnotesize Noise};
    \draw[->] (noise) -- (p1);
    \draw[->] (noise) -- (p2);
    \draw[->] (pF) -- (theta1);
    
    \draw[->] (pF2) -- (theta2);
    \draw[->] (theta2) -- (theta3);
    
    \node (loss) [below=2cm of noise] {$\mathcal{L}(\vtheta) = KL \left[ p_S \Vert p_R \right]$};
\end{tikzpicture}
}
\caption{Training process for Bayesian flow networks.}\label{fig:bfn_train}
\end{figure}
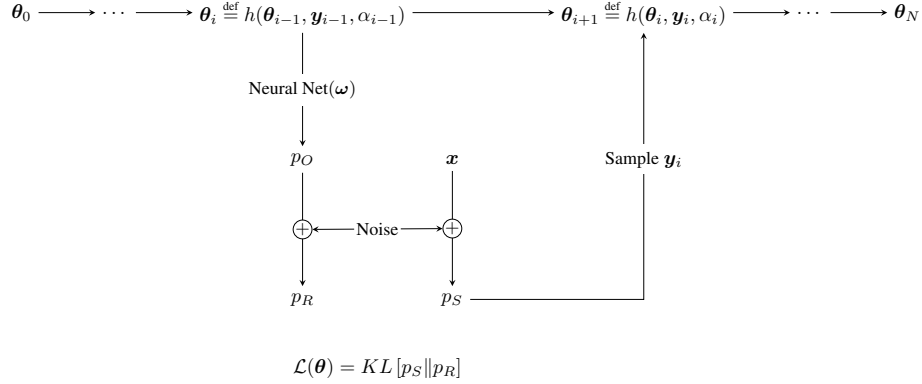

Figure~\ref{fig:bfn_train} represents the training process of Bayesian Flow Networks. The aim is to iteratively update the parameters of a distribution, \(\vtheta\), beginning with a prior distribution \(\vtheta_0\), so that eventually, sampling once from this distribution mirrors sampling from the data it is trained on.
In this process, accuracy \(\alpha_i\) refers to how well the updated parameters
 reflect the true data after observing information. The accuracy quantifies the expected quality of each update, meaning how much closer, in expectation, the updated distribution is to the true data distribution.

This training process involves the following key steps:
\begin{enumerate}
	\item \textbf{Generate \(\vtheta_{i}\):} Given a datapoint \(\vx\), the parameters \(\vtheta_{0}\) can be updated \(i\) times with noisy data to create parameters \(\vtheta_i\). We will later see that accuracies are additive and that \(\vtheta_{i}\) can be generated in a single step with accuracy \(\sum_{j=0}^{i-1} \alpha_j\).
	\item \textbf{Neural network transformation:} The parameters \(\vtheta_i\) are passed through a neural network, with weights \(\vomega\), which outputs the parameters of a new distribution. This is referred to as the \emph{output distribution} \(p_O\).
	\item \textbf{Addition of noise:} A new \emph{sender distribution} \(p_S\), is created by adding noise to the data according to a predefined schedule. Meanwhile, a \emph{receiver distribution} \(p_R\), is created by convolving the output distribution with the same noise.
	\item \textbf{KL divergence minimization:} The loss function is the KL divergence from the sender distribution \(p_S\) to the receiver distribution \(p_R\). The neural network weights, \(\vomega\), are updated by stochastic gradient descent to minimize this loss.
\end{enumerate}

This training procedure shows that if no noise is added to \(p_O\), the network will learn to collapse \(p_O\) onto \(\vx\). With noise added, however, the network learns to give a probability to all \(\vx^\prime\) relative to how likely they were to produce \(p_R\). The level of noise added is determined by an accuracy schedule, \(\{\alpha_0, \dots, \alpha_{N-1}\}\). The accuracy starts low and increases over time.

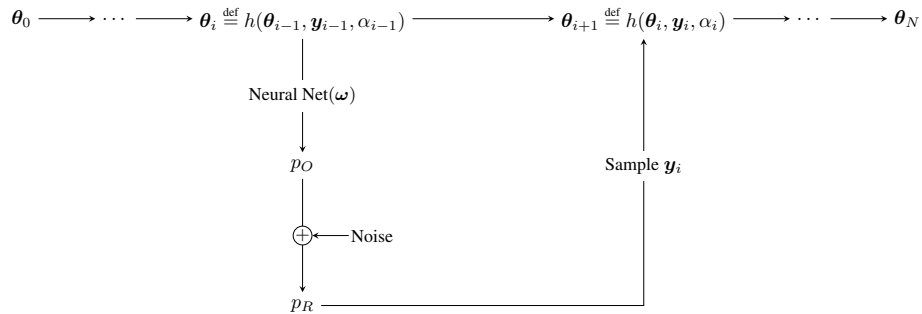
\begin{figure}[htbp]
	\center
    \scalebox{.75}{
	\begin{tikzpicture}[>=stealth, node distance=2cm, auto]

    \node (theta1) at (0,0) {$\vtheta_{i} \defeq h(\vtheta_{i-1}, \vy_{i-1}, \alpha_{i-1})$};
    
    \node (pF) [left=1cm of theta1] {$\dots$};
    \node (theta0) [left=1cm of pF] {$\vtheta_{0}$};
    \node (pO) [below=2cm of theta1] {$p_O$};
    \node (pR) [below=2cm of pO] {$p_R$};
    \node (pF2) [right=2.5cm of theta1] {$\vtheta_{i+1} \defeq h(\vtheta_i, \vy_i, \alpha_i)$};
    \node (theta2) [right=1cm of pF2] {$\dots$};
    \node (theta3) [right=1cm of theta2] {$\vtheta_{N}$};

    \draw[->] (theta1) to (pF2);
	\draw[->] (theta0) -- (pF);
    
   	\node (pS) [right=2cm of pR] {};
   	\node (x) at (pS|-pO) {};
   	
   	\draw[->, white] (x) -- node[midway, anchor=center, fill=white, sloped, circle, draw, inner sep=0pt, name=p2] {\small$+$} (pS);

    \draw[->] (theta1) -- node[midway, name=nn, anchor=center, fill=white] {\footnotesize Neural Net$(\vomega)$} (pO);
    \draw[->] (pO) -- node[midway, anchor=center, fill=white, sloped, circle, draw, inner sep=0pt, name=p1] {\small$+$} (pR);
    
    \draw[->] (pR) -| (pF2);
    \node (sample)[fill=white] at (x-|pF2) {{\footnotesize Sample} $\vy_i$};
    
    \node (noise)[inner sep=1] at ($(p1)!0.5!(p2)$) {\footnotesize Noise};
    \draw[->] (noise) -- (p1);
    \draw[->] (pF) -- (theta1);
    
    \draw[->] (pF2) -- (theta2);
    \draw[->] (theta2) -- (theta3);
    \end{tikzpicture}
    }
\caption{Generative process for Bayesian flow networks.}\label{fig:bfn_gen}
\end{figure}

Figure~\ref{fig:bfn_gen} represents the generative process of Bayesian flow networks. The key difference from the training process is that the parameters are updated based on samples from the receiver distribution, rather than the sender distribution. It unfolds as follows:

\begin{enumerate}
	\item \textbf{Initial parameterization:} The process starts with parameters, \(\vtheta_0\), of a prior distribution. For discrete data, these represent uniform probability.
	\item \textbf{Neural network transformation:} Similar to the training process, the neural network transforms the parameters \(\vtheta_i\) to produce an output distribution \(p_O\).
	\item \textbf{Noise injection:} Noise is convolved with the output distribution to create the receiver distribution.
	\item \textbf{Bayesian update:} A sample from the receiver distribution is used to update \(\vtheta_i\) using the Bayesian update function \(h\).
	\item \textbf{Iterate:} Step 2-4 is repeated \(N\) times, after which the parameters are fed into the neural network one last time to produce the final \(p_O\), from which a sample is taken. Note that the noise added to \(p_O\) to produce \(p_R\) follows the same accuracy schedule as during training.
\end{enumerate}

A comprehensive description of Bayesian Flow Networks is beyond the scope of this paper, but we aim to give the reader a clear understanding of how they differ from diffusion models. We will now look at what \(\vtheta\), \(h\), \(p_O\), \(p_R\), and \(p_S\) shown in Figure~\ref{fig:bfn_train} look like for categorical distributions.

Consider data represented as a \(D\) dimensional vector \(\Vx = \left(x^{(1)}, \dots, x^{(D)} \right) \in \left\{ 1, \dots, A \right\}^D\), where \(A\) is the maximum size of the state space over \(\vx^{(d)}\), and \(\left\{1, \dots, A \right\}\) is the set of integers from 1 and \(A\). We will model this as a categorical distribution. 

\paragraph{Input distribution}
The input distribution defines the probability of the data given the parameters fed into the neural network, as shown in Figures~\ref{fig:bfn_train} and \ref{fig:bfn_gen}. For discrete data, this distribution is modeled as a factorized categorical distribution with parameters \(\vtheta = \left( \vtheta^{(1)}, \dots, \vtheta^{(D)} \right)\), where each \(\vtheta^{(d)}\) includes parameters for a categorical distribution over \(A\) categories corresponding to variable \(d\). Specifically, \(\theta_a^{(d)}\) represents the probability of category \(a\) for the variable \(d\):
\begin{equation*}
	p_I(\Vx \mid \vtheta) = \prod_{d=1}^D p_I(\Vx^{(d)}\mid\vtheta^{(d)}).
\end{equation*}

\noindent
Initially, the input distribution is uniform, meaning \(\vtheta_0 = \left[\frac{1}{A}, \dots, \frac{1}{A}\right]\).

\paragraph{Output distribution} \(\Psi_{\vomega}(\vtheta_i, i)\) is a neural network model that takes as input a \(D\)-dimensional parameter vector \(\vtheta_i\), where each element are parameters of a categorical distribution. The output is of the same type. The output distribution for discrete data is defined based on the data \(\Vx\), model inputs \(\vtheta_i\), step counter \(i\), and resulting model outputs \(\Psi_{\vomega} \left( \vtheta_i, i \right) = \left( \Psi_{\vomega}^{(1)}\left( \vtheta_i, i \right), \dots, \Psi_{\vomega}^{(D)}\left( \vtheta_i, i \right) \right) \in \R^{A \times D}\). The network inputs 
\(\vtheta_i\) represents the parameters of the factorized categorical distribution \(p_I(\Vx \mid \vtheta_i)\), while \(i\) serves as an additional input that represents the process time. The output distribution is thus defined as 
\begin{equation*}
p_O(\Vx \mid \vtheta_i, i) = \prod_{d=1}^D \Psi^{(d)}_{\vomega}(\vtheta_i, i).
\end{equation*}

\noindent
Here, \(\Psi_{\vomega}^{(d)}(\vtheta_i, i)\) denotes \(A\) components of the network output corresponding to the parameters \(\left(\theta^{(d)}_1, \dots, \theta^{(d)}_A \right)\) of the categorical distribution for the \(d\)-th observation.

\paragraph{Sender distribution}
A sample from the sender distribution is used to update the parameters \(\vtheta\).
For \(\Vy = \left( y^{(1)}, \dots, y^{(D)} \right) \in \mathcal{Y}^D\), the sender distribution is defined as
\begin{equation*}
p_S \left( \Vy \mid \Vx; \alpha \right) = \N \left( \Vy \mid \alpha \left( A \Vex  - \bm{1} \right), \alpha A \bm{I} \right),
\end{equation*}
where \(\Vex\) is a unit vector of length \(A\) and element \(\Vx\) is \(1\), also known as a one-hot-encoding.
The accuracy of these samples is controlled by an accuracy parameter \(\alpha \in \R^+\).
When \(\alpha\) is low, the samples provide limited information about \(\Vx\). As \(\alpha\) increases, the samples become increasingly informative about \(\Vx\). Note that the value of \(\vy_i\) determines the amount of information, not the variance. The reason for this is that \(\vy_i\) will play the role of the logits. The value of \(\E(\vy_i)\) increases with \(\alpha_i\), which means that we get a higher focus on the values where \(\Vex = 1\).

\paragraph{Receiver distribution}
The receiver distribution is defined according to the output distribution \(p_O\), and \(p_S\), this takes the form
\begin{equation*}
    p_R(\Vy \mid \vtheta_i; i, \alpha) = \E_{p_O \left( \Vx^{\prime} \mid \vtheta_i; i \right)} \left[ p_S \left( \Vy \vert \Vx^\prime; \alpha \right) \right],
\end{equation*}

\noindent
In essence, this integrates over all \(\Vx^\prime \in \{1, \dots, A\}^D \), considering the contribution of each possible \(\Vx^\prime\) as weighted by its likelihood under the output distribution \(p_O(\Vx | \vtheta_i, i)\), effectively combines all potential sender distributions into a single receiver distribution.

The Bayesian update function for discrete data introduced in Figure~\ref{fig:bfn} is given by
\begin{equation*}
	h \left(\vtheta_{i}, \Vy_i \right) = \frac{e^{\Vy_i} \odot \vtheta_{i}}{\sum_{a=1}^A{e^{\Vy_{i,a}}\left(\vtheta_{i}\right)_a}},
\end{equation*}
where \(\odot\) refers to the Hadamard product. For a detailed derivation of the update function, refer to the original work by \citet{graves2024bayesian}.

At each step, the objective is to minimize the KL divergence from the sender distribution to the receiver distribution over all variables. As the noise in \(p_S\) approaches zero, the process is driven to the distribution that maximizes the likelihood of sampling data from the distribution \(p_O\).
\citet{graves2024bayesian} show that the loss function for an \(n\)-step procedure at step \(i\) is:

\begin{equation*}
\begin{split}
	L^N(\vomega; \mathbf{\Vx}, \Vy_i, \vtheta_i, i) = N \, \mathbb{E} \Bigg[ \ln \mathcal{N} \left( \Vy_i \mid \alpha_i (A\Vex - 1), \alpha_i A I \right) \\
-  \ln \left( \sum_{a=1}^A p_O \left( a \mid \vtheta_i, i \right) \mathcal{N} \left( \Vy_i \mid \alpha_i (A\Vea - 1), \alpha_i A I \right) \right) \Bigg].
\end{split}
\end{equation*}

Furthermore, \citet{graves2024bayesian} shows that if you let \(N\) go to infinity you get a continuous-time loss function. In that case, the generative process is not bound by \(N\) used during training. We omit the detailed derivation here, but \citet{graves2024bayesian} show that the continuous-time loss works marginally better than the discrete-time loss when 
$N$ is large. We will use the continuous-time loss in our experiments.
	
	\begin{algorithm}
		\caption{Generating Discrete Samples With Bayesian Flow Networks}\label{alg:gen}
		\begin{algorithmic}[1]
			\Require $\beta(1) \in \mathbb{R}^+$, number of steps $n \in \mathbb{N}$, number of categories $K \in \mathbb{N}$
			\State $\vtheta \gets \left( \frac{\bm{1}}{\bm{K}} \right)$
			\For{$i = 1 \text{ to } n$}
				
				\State $\bm{k} \sim p_O(\cdot \mid \vtheta, i)$
				\State $\alpha \gets \beta(1)(\frac{2i-1}{n^2})$
				\State $\bm{y} \sim \mathcal{N}(\alpha (K \bm{e_k} - 1), \alpha K \mathbf{I})$
				\State $\vtheta^\prime \gets e^{\bm{y}} \odot \vtheta$
				\State $\vtheta \gets \frac{\theta^\prime}{\sum_k \theta^\prime_k}$
			\EndFor
			\State $\bm{k} \sim p_O(\cdot \mid \vtheta, 1)$
		\end{algorithmic}
	\end{algorithm}

\section{Related Work}\label{related-work}
Several recent works have considered offline RL as a sequence modeling problem.
Here we will consider the two most prominent. 

\subsection{Diffuser}\label{diffuser}
Diffuser~\citep{janner2022diffuser} models state-action trajectories using an unconditional diffusion model. At test time, the current state is imposed through inpainting, while gradients from a differentiable reward model guide sampling toward high-return trajectories. Terminal states can likewise be imposed through inpainting to obtain goal-conditioned plans.
The authors further show that the inpainting technique can be used to condition desired end states, effectively allowing the model to solve planning problems it has not been specifically trained for.

\subsection{Decision Diffuser}\label{decision-diffuser}
The Decision Diffuser~\cite{ajay2022conditional} differs from Diffuser in two main ways:  how it models actions, and how it conditions on rewards. 
First, Decision Diffuser leverages an inverse dynamics model to capture the relationship between states and actions. 
This inverse dynamics model estimates actions conditioned on states, effectively predicting the action that brought the environment from one state to the next. 
This lets the diffusion model focus only on state sequences, rather than on sequences containing both state and action. 
The authors show empirically that using an inverse dynamics model is advantageous in deterministic environments, but that the performance reduces to the same level as the Diffuser as more stochasticity is introduced into the environment. 
Second, the Decision Diffuser conditions on return-to-go in a classifier-free manner. 
This means that the required return is fed into the model during training so that the model learns which sequences to associate with that return. 
The desired return can again be fed into the model at test time when generating a sequence of future states.

Native categorical diffusion models provide an alternative route to discrete generation \citep{austin2023structureddenoisingdiffusionmodels,lou2024discretediffusionmodelingestimating}. In the discrete experiments below, we therefore compare the categorical BFN with a uniform-categorical diffusion model that predicts the clean trajectory and uses the exact reverse posterior.

\section{Method}\label{method}
We propose a sequence-generating approach to reinforcement learning
based on Bayesian flow networks capable of planning in both discrete and
continuous domains. From now on we will refer to our method as \lfm{}. 
Like the Decision Diffuser~\citep{ajay2022conditional}, \lfm{} only models sequences of \textit{states}. Return is supplied as a network condition, whereas the current state is fixed by inpainting during sampling.
We then utilize a second inverse-dynamics network to model the actions conditioned on the states.
In the context of diffusion models, this approach has been shown to provide superior performance compared to modeling state-action pairs from a single model~\citep{janner2022diffuser}. 
We expect the same benefits when using BFNs as the generative model, but consider further examination of this hypothesis as future work.

In our method, the BFN is trained to generate
sequences conditioned on the \textit{return}. 
The return is the sum of all discounted future rewards and is
therefore a measure of the quality of a sequence. 
The network learns to model the distribution over sequences with both high and low returns. 
At test time, the desired return biases generation toward trajectories with the corresponding cumulative reward.

\subsection{Condition on Return}
There are two obvious ways we can condition on return: 
Either as done by the Diffuser~\citep{janner2022diffuser}  or as done by the Decision Diffuser~\citep{ajay2022conditional}. 
Considering that the latter option is significantly easier to implement, showed better performance, and does not involve training an extra model, we opted to condition directly on return in a classifier-free manner as was done in Decision Diffuser~\citep{ajay2022conditional}. 
At each BFN step, the neural network receives the current parameters of the state distribution, factorized over trajectory time, together with the return and BFN time. The observed current state is not supplied through a separate direct-conditioning pathway.

By conditioning directly on return during the generation process, the model learns to generate trajectories associated with different levels of cumulative reward. At inference time, this allows the generation process to be biased toward higher-return trajectories by conditioning on a desired return value. Furthermore, adopting the Decision Diffuser-style conditioning framework enables straightforward integration with existing sequence-modeling architectures, including the temporal U-net architecture used by \citet{ajay2022conditional}.

\subsection{Conditioning on Current State}
When Diffuser~\citep{janner2022diffuser} and Decision Diffuser~\citep{ajay2022conditional} condition on the current state, 
they both apply an inpainting technique specific to diffusion models, see 
Eq.~\eqref{equ:inpainting_total}. 
During the reverse diffusion process, the part of the sequence that is known (the first state), is replaced by the true value diffused to the appropriate amount for that step. 
We have adopted a similar technique for BFNs. 
For discrete data, we set the probabilities of the categorical distribution of the known parts of the object to the appropriate probability for that BFN step. 
Algorithm~\ref{alg:inp_cond} shows this method implemented for discrete data. 
The alterations to BFN sampling (Algorithm~9 in \citet{graves2024bayesian}) are shown in Algorithm~\ref{alg:inp_cond}. For continuous data, the Bayesian update at each step is made similarly, see Algorithm \ref{alg:inp_cond_cont} in the Appendix. 

The operations involving the mask and conditioning values are those added to the regular BFN algorithm. 
In Algorithm~\ref{alg:inp_cond}, the mask \(\bm{m}\) selects the observed variables, and \(\bm{c}\) supplies their clean categorical values.
All reported experiments use this inpainting mechanism to fix the current state throughout sampling.

For the categorical experiments, we retain the quadratic BFN accuracy schedule \(\beta(t)=\beta(1)t^2\), but parameterize its endpoint by \(C=K\beta(1)\), where \(K\) is the number of categories. This keeps the terminal correct-versus-incorrect message-mean separation from changing merely because a representation uses more categories. We use \(C=36\), 18, and 72 for Empty-Random, DoorKey, and BlockedUnlockPickup, respectively.
The continuous-time BFN objective is augmented by clean-data cross entropy with weight 0.5. Reported BFN results use the categorical sampling procedure in Algorithm~\ref{alg:inp_cond}, with 12 updates for Empty-Random and 16 for DoorKey and BlockedUnlockPickup.
For categorical tasks, we optimize \(\mathcal{L}_{\mathrm{BFN}}+0.5\mathcal{L}_{\mathrm{CE}}\), where \(\mathcal{L}_{\mathrm{CE}}\) is the cross-entropy for predicting the clean category from the BFN input parameters.

\begin{algorithm}[H]
  \caption{Inpainting Conditioning for Discrete Random Variables}\label{alg:inp_cond}
  \begin{algorithmic}[1]
\Require \(\beta(1) \in \R^+\), number of steps \(n \in \mathbb{N}\), number of categories \(K\), \textcolor{black}{mask \(\bm{m}\), condition \(\bm{c}\)}
\State \(\vtheta \gets \frac{\bm{1}}{\bm{K}}\)
\For{\(i=1 \text{ to } n\)}
    \State \(t \gets \frac{i-1}{n}\)
    \State \(\mathbf{k} \sim\) \Call{discrete\_output\_distribution}{$\vtheta, t$}
    \State \(\alpha \gets \beta(1) \left( \frac{2i-1}{n^2} \right)\)
    \State \(\mathbf{y} \sim \N \left(\alpha (K \mathbf{e_k} - \bm{1}), \alpha K \I \right)\)
    \State {\color{black} \(\mathbf{y_c} \gets \alpha (K \mathbf{e_c} - \bm{1})\)}
    \State {\color{black} \(\mathbf{y} \gets \bm{m} \odot \mathbf{y_c} + (1 - \bm{m}) \odot \mathbf{y} \)}
    \State \(\vtheta^\prime \gets e^\mathbf{y} \odot \vtheta\)
    \State \(\vtheta = \frac{\vtheta^\prime}{\sum_k \vtheta^\prime_k}\)
\EndFor
\State \(\mathbf{k} \sim\) \Call{discrete\_output\_distribution}{$\vtheta, 1$}
\end{algorithmic}
\end{algorithm}

\section{Experiments}\label{experiments}
We evaluate our method on two sets of tasks, one with discrete
action and state space, and one with continuous action and state space.
In the discrete case, we use a grid world environment. For the continuous
case, we use the D4RL~\citep{fu2020d4rl} datasets with the Gym-Mujoco suite of
environments. We compare our method to the Decision
Diffuser~\citep{ajay2022conditional} and other state of the art offline
RL methods.
The discrete experiments test whether a categorical BFN can generate executable state plans without hand-coded transition rules. The continuous experiments test whether the same planning formulation remains competitive on standard offline RL benchmarks.

\subsection{Discrete Experiments}\label{gridworld}
We evaluate BFN-RL on three MiniGrid tasks, as well as FrozenLake and Sokoban. In each environment, a categorical BFN generates state sequences and a learned inverse-dynamics model maps consecutive states to actions from the environment's native categorical action space. The MiniGrid environments have seven possible actions. We use the same temporal U-Net family across all discrete tasks, with task-specific widths, planning horizons, state representations, and conditioning inputs.

Empty-Random requires navigation to a randomly located goal. DoorKey additionally requires the agent to collect a key and open a door, while BlockedUnlockPickup requires moving an obstructing object before collecting a target box. Their training sets contain 2,000, 5,000, and 3,000 successful trajectories, respectively. The DoorKey and BlockedUnlockPickup datasets include random perturbations followed by expert recovery. We additionally evaluate slippery FrozenLake-8$\times$8 navigation and two-box Sokoban-7$\times$7 using 30,000 and 10,000 successful trajectories, respectively.

At each replanning step, the observed state is fixed through inpainting and one trajectory is sampled. Empty-Random uses a planning horizon of 16 and replans after every action. DoorKey and BlockedUnlockPickup use a horizon of 32 and execute at most 16 actions from each sampled trajectory, replanning earlier if the observed state departs from the generated plan. FrozenLake and Sokoban use horizons of 64 and 32, respectively, and replan after every action. No hand-coded transition projection or analytic inverse dynamics is used.

The MiniGrid experiments condition generation on return, whereas FrozenLake and Sokoban use goal conditioning. To isolate the effect of the generative model, the categorical-diffusion baseline is matched to BFN-RL in training data, temporal U-Net architecture, state representation, conditioning signal, current-state inpainting, inverse-dynamics model, planning horizon, action-execution protocol, and evaluation seeds. The baseline uses uniform categorical corruption with a cosine signal-survival schedule.

\begin{table}[ht]
\centering
\small
\begin{tabular}{llcc}
\toprule
Task & Generator & Success (\%) & Mean return \\
\midrule
Empty-Random-6$\times$6 & Categorical diffusion & \textbf{100.0 $\pm$ 0.0} & \textbf{0.941 $\pm$ 0.012} \\
 & \lfm{} & 99.7 $\pm$ 0.6 & 0.895 $\pm$ 0.004 \\
\midrule
DoorKey-8$\times$8 & Categorical diffusion & 99.3 $\pm$ 0.6 & 0.916 $\pm$ 0.010 \\
 & \lfm{} & \textbf{100.0 $\pm$ 0.0} & \textbf{0.934 $\pm$ 0.002} \\
\midrule
BlockedUnlockPickup & Categorical diffusion & 37.0 $\pm$ 13.5 & 0.305 $\pm$ 0.113 \\
 & \lfm{} & 32.7 $\pm$ 8.3 & 0.279 $\pm$ 0.072 \\
\specialrule{1.2pt}{0.9ex}{0.7ex}
FrozenLake-8$\times$8 & Categorical diffusion & 85.5 $\pm$ 1.0 & \\
 & BFN-RL & \textbf{86.7 $\pm$ 1.3} & \\
\midrule
Sokoban-7$\times$7 & Categorical diffusion & 59.0 $\pm$ 2.6 & \\
 & BFN-RL & \textbf{68.7 $\pm$ 2.5} & \\
\bottomrule
\end{tabular}
\caption{Matched discrete-generator comparison. Entries are mean $\pm$ sample standard deviation across three training seeds. Evaluation uses 100 episodes per seed except for FrozenLake, which uses 200. Environment and planner seeds are matched between generators.}
\label{tab:minigrid-main}
\end{table}

The diffusion baseline reaches the goal more quickly in Empty-Random, while BFN-RL has the higher return in DoorKey. On BlockedUnlockPickup, variation between training seeds is larger than the difference in mean success. Overall, BFN-RL is a competitive categorical planner, although it does not uniformly outperform categorical diffusion.

BFN-RL and categorical diffusion perform similarly on FrozenLake. On Sokoban, BFN-RL has higher success for each of the three training seeds and improves the mean success rate by 9.7 percentage points. This identifies a harder discrete task on which the BFN generator provides a consistent advantage under the matched protocol.

\subsection{Continuous Control}\label{continuous-control}
The original Gaussian formulations of Diffuser~\citep{janner2022diffuser} and Decision Diffuser~\citep{ajay2022conditional} are not natively categorical. We evaluate BFN-RL on D4RL MuJoCo to test whether the same planning framework remains competitive in continuous control.

Table~\ref{tab:cont_res} shows the performance of \lfm{} compared to state-of-the-art algorithms. 
The table shows that \lfm{} is competitive on most datasets.

\begin{table}[H]
	\centering
	\small
	\begin{tabular}{llccccccccc}\toprule
		\textbf{Dataset} & \textbf{Environment} & \textbf{BC} & \textbf{CQL} & \textbf{IQL} & \textbf{DT} & \textbf{TT} & \textbf{MOReL} & \textbf{Diffuser} & \textbf{DD} & \textbf{\lfm}\\\midrule
		Med-Expert & HalfCheetah & 55.2  & 91.6  & 86.7  & 86.8  & {95}    & 53.3  & 79.8  & 90.6 & 97.2 $\pm$ 0.1\\
		Med-Expert & Hopper 	 & 52.5 	 & 105.4 & 91.5  & 107.6 & {110.0} & 108.7 	& 107.2 & {111.8} & 110.3 $\pm$ 0.8 \\
		Med-Expert & Walker2d 	 & {107.5} & {108.8} & {109.6} & {108.1} & 101.9 & 95.6  & {108.4} & {108.8} & 106.6 $\pm$ 4.9 \\
		\midrule
		Medium & HalfCheetah & 42.6  & 44.0  & 47.4  & 42.6  & 46.9  & 42.1  & 44.2  & {49.1} & 47.8 $\pm$ 0.1 \\
		Medium & Hopper 	 & 52.9 	 & 58.5 & 66.3  & 67.6 & 61.1 & {95.4} & 58.5 & 79.3 & 72.2 $\pm$ 7.1 \\
		Medium & Walker2d 	 & 75.3  & 72.5 & 78.3 & 74.0 & 79 & 77.8 & 79.7 & {82.5} & 66.9 $\pm$ 5.5 \\
		\midrule
		Med-Replay & HalfCheetah & 36.6  & {45.5}  & {44.2}  & 36.6  & 41.9  & 40.2  & 42.2  & 39.3 & 42.5 $\pm$ 0.5\\
		Med-Replay & Hopper 	 & 18.1	 & 95 & 94.7  & 82.7 & 91.5 & 93.6 & 96.8 & {100} & 91.2 $\pm$ 6.3\\
		Med-Replay & Walker2d 	 & 26.0  & 77.2 & 73.9 & 66.6 & {82.6} & 49.8 & 61.2 & 75 & 58.5 $\pm$ 4.8 \\
		\midrule
		\multicolumn{2}{c}{\textbf{Average}} & 51.9  & 77.6 & 77 & 74.7 & 78.9 & 72.9 & 75.3 & {81.8} & 77.0 \\
	\bottomrule
	\end{tabular}
	\caption{
    The table summarizes the test performance of \lfm{} and various other methods for continuous control. 
    The results indicate that \lfm{} can perform at a level comparable to the state of the art. 
    We report mean and standard error over 3 random seeds. All numbers except for \lfm{} are from \cite{ajay2022conditional}.} \label{tab:cont_res}
\end{table}

\section{Conclusion}
In this work, we introduced a novel approach to reinforcement learning that leverages Bayesian flow networks~\citep{graves2024bayesian} for sequence generation. Our method is capable of planning in both discrete and continuous domains. This shared planning framework also suggests a natural extension to joint continuous-categorical environments, in which each trajectory variable is assigned the corresponding continuous or categorical BFN distribution within one model. The current state is fixed by inpainting throughout sampling, while return conditioning avoids the need for an additional return classifier.

Across the discrete tasks, BFN-RL is competitive with matched categorical diffusion and is consistently stronger on Sokoban. In continuous control, it remains competitive with established offline trajectory models. Together, these results support BFNs as a common planning framework for categorical and continuous domains.

Independent work on Guided-BFNs reports strong results in continuous trajectory-planning environments by supplementing conditional guidance with gradients from a learned reward model during sampling~\citep{li2025guided}. That work considers continuous state--action trajectories, whereas our categorical experiments address planning with discrete states and actions.

\section*{Declarations}
This research was funded by internal funding from the Norwegian University of Science and Technology.

\begin{appendices}

\section{Hyperparameters}\label{secA1}

Here we present hyperparameters used in the experiments. For the discrete experiments, we used:
\begin{itemize}
	\item Quadratic accuracy schedule \(\beta(t)=\beta(1)t^2\), with \(K\beta(1)=36\), 18, and 72 for Empty-Random, DoorKey, and BlockedUnlockPickup, respectively. The planning horizons are 16, 32, and 32.
	\item Learning rate \(3\times10^{-4}\), batch size 128, Adam optimizer~\citep{Kingma2014AdamAM}, linear warmup for 200 updates on Empty-Random and 300 updates otherwise, followed by cosine decay.
	\item A generic temporal U-Net with base width 32 for Empty-Random, 48 for DoorKey, and 48 or 96 in the BlockedUnlockPickup sweep.
	\item A learned inverse-dynamics MLP with two hidden layers of 256 units.
	\item The continuous-time categorical BFN objective is augmented by a cross-entropy term with weight 0.5.
	\item Categorical BFN sampling uses 12 updates for Empty-Random and 16 for DoorKey and BlockedUnlockPickup, temperature 0.7 for Empty-Random and 0.9 otherwise, and one generated candidate.
	\item Empty-Random replans after each action. For DoorKey and BlockedUnlockPickup, at most 16 actions are executed from a plan, with immediate replanning after a state disagreement.
	\item The matched categorical-diffusion baseline uses cosine uniform corruption, clean-state cross entropy, and 12 or 16 deterministic reverse updates, matching the BFN update count for each task.
	\item FrozenLake and Sokoban use horizons of 64 and 32, base widths of 64 and 96, batch sizes of 256 and 64, and $K\beta(1)=72$. Both use cross-entropy weight 0.5, temperature 0.9, 16 sampling updates, and one-action replanning.
\end{itemize}

Most hyperparameters and model architectures for continuous experiments that are not specific to
Bayesian Flow Networks are similar to those used in the official
Decision Diffuser implementation. 

Hyperparameters used for the continuous experiments:

\begin{itemize}
	\item The Inverse dynamics model is an MLP with two layers with 256 units and ReLU activations.
	\item \(\epsilon_\theta\) and \(f_\phi\) are trained for \(2 \times 10^6\) steps using the Adam optimiser~\citep{Kingma2014AdamAM} with a batch size of 256,  a learning rate of \(5 \times 10^{-5}\).
	\item We use a planning horizon \(H\) of 20.
	\item For testing we used an exponential moving average of the weights with decay \(\alpha=0.9995\)
	\item First-order BFN solver of \citet{xue2024unifying} using 10 sampling steps.
    \item Guidance weight 1.2.
    \item Temperature 0.5.
	\item \(\sigma_1 = 0.01\).
\end{itemize}

\section{Algorithms}
Algorithm~\ref{alg:inp_cond_cont} gives the continuous-variable counterpart of the inpainting procedure in Algorithm~\ref{alg:inp_cond}.

\begin{algorithm}[H]
      \caption{Inpainting Conditioning for Continuous Random Variables}\label{alg:inp_cond_cont}
      \begin{algorithmic}[1]
	\Require \(\sigma_1 \in \R^+\), number of steps \(n \in \mathbb{N}\), \textcolor{black}{mask \(\bm{m}\), condition \(\bm{c}\)}
	\State \(\vmu \gets \bm{0}\)
	\State \(\rho \gets 1\)
	\For{\(i=1 \text{ to } n\)}
		\State \(t \gets \frac{i-1}{n}\)
		\State \(\mathbf{\hat{x}}(\vtheta, t) \gets\) \Call{cts\_output\_distribution}{$\vmu, t, 1-\sigma_1^2$}
		\State \(\alpha \gets \sigma_1^{-2i/n} \left( 1 - \sigma_1^{2/n} \right)\)
		\State \(\mathbf{y} \sim \N \left(\mathbf{\hat{x}}(\vtheta, t), \alpha^{-1} \I \right)\)
		\State {\color{black} \(\mathbf{y_c} \gets (1 - \sigma_1^{2t}) \bm{c}\)}
		\State {\color{black} \(\mathbf{y} \gets \bm{m} \odot \mathbf{y_c} +(\bm{1} - \bm{m}) \odot \mathbf{y} \)}
		\State \(\vmu \gets \frac{\rho \vmu + \alpha \mathbf{y}}{\rho + \alpha}\)
		\State \(\rho \gets \rho + \alpha\)
	\EndFor
	\State \(\bm{\hat{x}}(\vtheta, 1) \gets\) \Call{cts\_output\_distribution}{$\vmu, 1, 1-\sigma_1^2$}
\end{algorithmic}
\end{algorithm}

Here \(\rho\) is the precision of the continuous BFN input distribution; it starts at the unit-precision prior and accumulates the message precisions \(\alpha\).

\end{appendices}

\bibliographystyle{plainnat} 
\bibliography{sn-bibliography}

\end{document}